\documentclass[conference]{IEEEtran}
\IEEEoverridecommandlockouts
\usepackage{cite}
\usepackage{amsmath,amssymb,amsfonts,bm}
\usepackage{algorithmic}
\usepackage{graphicx}
\usepackage{textcomp}
\usepackage{xcolor}
\usepackage{amsmath,lipsum}
\usepackage[ruled]{algorithm2e}
\usepackage{graphicx}
\usepackage{float}
\usepackage{subfigure}
\usepackage{multirow}
\usepackage{multicol}
\usepackage{arydshln}
\usepackage{cite}
\usepackage{stfloats}
\usepackage{booktabs}
\usepackage{caption}
\usepackage{makecell}
\usepackage{graphicx}
\usepackage{textcomp}
\usepackage{array} 
\usepackage{longtable}
\usepackage{booktabs}
\usepackage{float}

\def\BibTeX{{\rm B\kern-.05em{\sc i\kern-.025em b}\kern-.08em
    T\kern-.1667em\lower.7ex\hbox{E}\kern-.125emX}}
\begin{document}

\title{Spatiotemporal Attention-based Semantic Compression for Real-time Video Recognition   \\ }
\author{
\IEEEauthorblockN{
Nan Li\IEEEauthorrefmark{1},
Mehdi Bennis\IEEEauthorrefmark{2},
Alexandros Iosifidis\IEEEauthorrefmark{1}, and
Qi Zhang\IEEEauthorrefmark{1}}
\IEEEauthorblockA{\IEEEauthorrefmark{1}DIGIT, Department of Electrical and Computer Engineering, Aarhus University, Denmark}
\IEEEauthorblockA{\IEEEauthorrefmark{2}Centre for Wireless Communications, University of Oulu, Finland}
\IEEEauthorblockA{Emails: {linan}@ece.au.dk,\
{mehdi.bennis}@oulu.fi,\
    \{{ai, qz}\}@ece.au.dk
}}

\maketitle

\begin{abstract}
This paper studies the computational offloading of video action recognition in edge computing.
To achieve effective semantic information extraction and compression, following semantic communication we propose a novel spatiotemporal attention-based autoencoder (STAE) architecture, including a frame attention module and a spatial attention module,  to evaluate the importance of frames and pixels in each frame. Additionally, we use entropy encoding to remove statistical redundancy in the compressed data to further reduce communication overhead. At the receiver, we develop a lightweight decoder that leverages a 3D-2D CNN combined architecture to reconstruct missing information by simultaneously learning temporal and spatial information from the received data to improve accuracy. To fasten convergence, we use a step-by-step approach to train the resulting STAE-based vision transformer (ViT\_STAE) models. Experimental results show that ViT\_STAE can compress the video dataset HMDB51 by $104 \times$ with only $5\%$ accuracy loss, outperforming the state-of-the-art baseline DeepISC. The proposed ViT\_STAE achieves faster inference and higher accuracy than the DeepISC-based ViT model under time-varying wireless channel, which highlights the effectiveness of STAE in guaranteeing higher accuracy under time constraints.
\end{abstract}


\section{Introduction}
Deep Neural Networks (DNNs) have revolutionized the landscape of IoT applications by achieving better decision-making and automation. Video action recognition, as a representative computer vision application, has diverse use cases ranging from video surveillance to autonomous driving and human-computer interaction. To achieve higher accuracy in video analysis, deeper DNNs, such as vision transformers (ViT) with massive multiply-accumulate operations \cite{tong2022videomae}, are often desired. However, the high computational and memory requirements of DNNs challenge \textit{local computing} on resource-constrained IoT devices.

To address this challenge, \textit{mobile edge computing} (MEC) is a promising approach that allows IoT devices to offload computational tasks directly to edge servers (ESs) in their proximity \cite{Nan2022}. However, the time-varying wireless channel can impact offload decision-making, as the fluctuation in communication time may result in inference missing the service deadline. The emerging paradigm of semantic communication is expected to enable effective communication between IoT devices and ESs. Unlike traditional content-agnostic communication systems, semantic communication focuses on the effectiveness of the semantic content in the source data, rather than the average information associated with the statistical characteristics of source data \cite{9955312}. In task-oriented semantic communication, the deep learning-based semantic encoder learns the latent representations of the source data to extract semantic information, and the semantic decoder uses the received semantic information to execute its intended task. For DNN-based applications, the effectiveness of semantic content is usually measured by inference accuracy. 

This paper proposes a novel spatiotemporal attention-based autoencoder (STAE) to efficiently reduce data size and communication overhead to meet the deadline while ensuring high inference accuracy for a deep video recognition task in MEC. Specifically, the proposed STAE uses spatiotemporal attention to identify the most important frames and pixels per frame, transmitting the most informative features.
\subsection{Prior Works}
To reduce computational overhead and memory footprint, existing works proposed various compression methods of ViT. Chavan et al. \cite{Chavan_2022_CVPR} proposed ViT-Slim, which prunes 40\% of unnecessary weights and achieves a $2.5 \times$ reduction in computation while maintaining inference accuracy. Li et al. \cite{li2022qvit} utilized 4-bit quantization to compress the ViT model up to $6.14 \times$ with only a $1\%$ decrease in accuracy. Wu et al. \cite{tiny_vit} used knowledge distillation to train a smaller student model, resulting in a $4.2 \times$ reduction in parameters with a slight drop in accuracy. However, the compressed ViT still has numerous parameters that make deployment on resource-constrained IoT devices challenging. As such, more attention has been paid to effective semantic information coding for reliable inference.

Motivated by task-oriented semantic communication, Shao et al. \cite{BottleNet++} proposed BottleNet++ to compress the feature by using a CNN module to resize the feature dimension. However, directly resizing feature dimensions may compromise the effective representation of the semantic information and result in accuracy degradation. Akbari et al. \cite{DeepSIC} proposed a deep semantic segmentation-based layered image compression. However, it pursues a rather conventional pixel accuracy but fails to recover semantic information. Huang et al. \cite{huang} proposed a deep learning-based image semantic coding (DeepISC) for feature
extraction and recovery. However, directly sending the latent representation, semantic segmentation and the discrepancy between the input and synthesized data instead of the most important semantic information may cause excessive communication overhead. 

\begin{figure*}[]
    \centering
\includegraphics[width=0.89\textwidth]{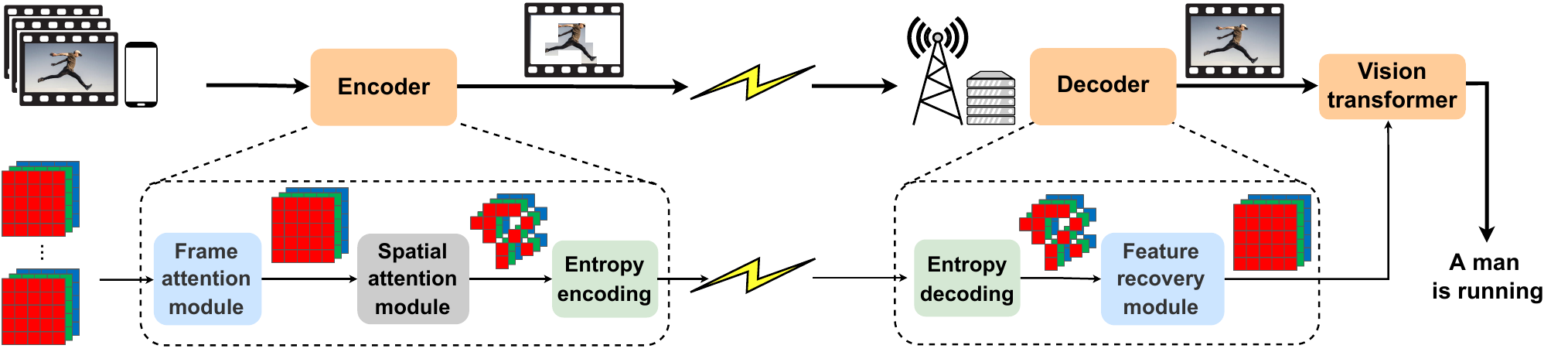}
    \caption{The proposed spatiotemporal attention-based autoencoder for video action recognition.}
        \vspace{-3mm}
    \label{fig:AECNN}
\end{figure*}
\subsection{Main Contributions}
This paper studies the computational offloading of video recognition task in an MEC system and proposes STAE for semantic information extraction and compression thereby meeting real time constraints while ensuring high inference accuracy. Our contributions include: 
\begin{itemize}
    \item We design a spatiotemporal attention module consisting of a \textit{frame attention} (FA) module and \textit{spatial attention} (SA) module to evaluate the importance of individual frames and pixels within each frame for inference, respectively. This approach allows for video compression by discarding less important frames and pixels, transmitting only the most important semantic information. Additionally, we exploit entropy encoding to eliminate statistical redundancies in the semantic information to further reduce communication overhead;
    \item We design a light-weight \textit{feature recovery} (FR) module that uses a 3D-2D CNN architecture to recover missing information by simultaneously learning the temporal structure and spatial features from the received semantic information, thereby improving inference accuracy;
    \item We train the resulting STAE-based ViT (ViT\_STAE) on video dataset HMDB51 \cite{HMDB} step-by-step to fasten convergence. Experimental results show that ViT\_STAE achieves $104 \times$ compression with only $5\%$ accuracy loss, which outperforms the state-of-the-art work, DeepISC \cite{huang}. Moreover,  ViT\_STAE achieves faster inference and higher accuracy than DeepISC-based ViT under different wireless channel conditions.
\end{itemize} 
\section{System Model and Problem Formulation}\label{section:system_model}
This paper considers the offloading of computation-intensive applications in an MEC network consisting of an IoT device and an ES, with an emphasis on video action recognition utilizing the state-of-the-art DNN model, ViT \cite{tong2022videomae}.
As shown in Fig. 1, a spatiotemporal attention-based autoencoder composed of a semantic encoder and a semantic decoder, is  employed to compress the visual data at a pre-defined frame and spatial budget, in order to recover the received data for inference. We use two sets $\mathcal{A}= \{\alpha_1, \alpha_2, \cdots \alpha_K \}$ and $\mathcal{B}= \{\beta_1, \beta_2, \cdots \beta_M \}$ to denote the pre-defined frame budget and spatial budget within each frame, where $\alpha_k $ represents the number of frames used for video recognition, and $\beta_m $ represents the percentage of pixels used from the selected $\alpha_k$ frames for video recognition. 

The complete execution of an inference task involves a sequence of stages. The IoT device first uses the FA module to evaluate the importance of each frame and prunes less important ones based on a frame budget $\alpha_k$. This step is followed by using the SA module to select important pixels of the selected $\alpha_k$ frames based on the spatial budget $\beta_m$. Subsequently, the remaining informative features are passed through an entropy encoding module to eliminate statistical redundancies in the data which are then transmitted to the ES over wireless channels. 
Finally, the ES decodes the received data using the entropy decoding module and uses the FR module to reconstruct missing information in each frame before applying the ViT model for inference and sending the feedback to the IoT device.
\subsection{Communication Time}
A video recognition task can be denoted as $X \in \mathbb{R}^{F \times C \times H \times W}$, where $F$, $C$, $H$, and $W$ correspond to the number of frames, channel, height, and width of a video, respectively. Note that channel $C$ refers to the Red, Green, and Blue (RGB) color channels of an original video, and $X$ is typically represented as a float32 type tensor in deep learning frameworks. Assuming the semantic encoder using frame budget, $\alpha_k$ and spatial budget, $\beta_m$, for the $\alpha_k$ frames, the size of a compressed video data in bytes is  
\begin{equation}
S_{\alpha_k, \beta_m} = 
    4 \cdot \alpha_k \cdot C \cdot \beta_m \cdot H \cdot W.
\end{equation}

Computation offloading involves transmitting the video data and its inference result between the IoT device and ES. Since the output of the ViT model is very small, usually a number or few values representing the classification or detection result, the feedback transmission delay is negligible. Therefore, the total communication time of transmitting a compressed video can be calculated as
\begin{equation}
t_{\alpha_k, \beta_m}^{\textit{com}} = 
    S_{\alpha_k, \beta_m}/\gamma,
\end{equation}
where $\gamma$ is the uplink transmission data rate from an IoT device to an ES.
\subsection{Computation Time}
The total computation time of an inference task can be expressed as the sum of the computation time for the FA and SA modules, the entropy encoding and decoding process, the FR module, and the inference time of the ViT model. Therefore, we denote the total computation time for a given frame budget $\alpha_k$ and spatial budget $\beta_m$ as $t_{\alpha_k, \beta_m}^{\textit{cmp}}$ and express it mathematically as,
\begin{equation}
\small
t_{\alpha_k, \beta_m}^{\textit{cmp}} = t_{\alpha_k}^{\textit{FA}} + t_{\alpha_k,\beta_m}^{\textit{SA}} + t_{\alpha_k,\beta_m}^{\textit{EE}} + t_{\alpha_k,\beta_m}^{\textit{ED}}+ t_{\alpha_k,\beta_m}^{\textit{FR}}+ t_{\alpha_k}^{\textit{ViT}},
\end{equation}
where $t_{\alpha_k}^{\textit{FA}}$, $t_{\alpha_k,\beta_m}^{\textit{SA}}$ and $t_{\alpha_k,\beta_m}^{\textit{EE}}$ represent the time taken by IoT device to perform the computation of FA module, spatial attention module and entropy encoding, respectively; $t_{\alpha_k,\beta_m}^{\textit{ED}}$, $t_{\alpha_k,\beta_m}^{\textit{FR}}$ and $t_{\alpha_k}^{\textit{ViT}}$ represent the time taken by ES to perform the entropy decoding and computation of FR module and ViT model, respectively.
\begin{figure*}
    \centering
    \includegraphics[width=0.96\textwidth]{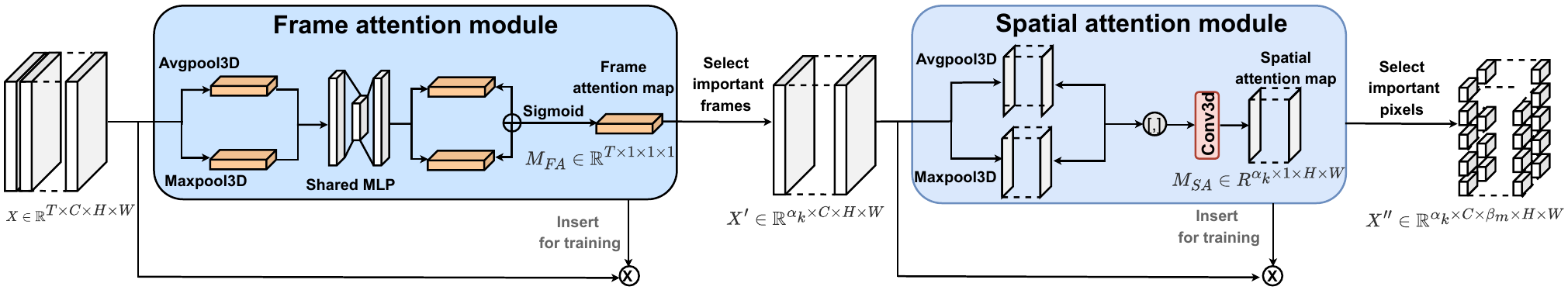}
    \caption{The proposed spatiotemporal attention architecture of the Encoder.}
        \vspace{-5mm}
    \label{fig:encoder}
\end{figure*}
\subsection{Completion Time}
The complete execution of an inference task involves data transmission and computation. Therefore, the duration required to complete an inference task, at the given frame budget $\alpha_k$ and spatial budget $\beta_m$, is given by
\begin{equation}
    \begin{array}{*{20}{c}}
         t_{\alpha_k,\beta_m} = t_{\alpha_k,\beta_m}^{\textit{com}} + t_{\alpha_k,\beta_m}^{\textit{cmp}}, &\forall  \alpha_k \in \mathcal{A} \ \textrm{and} \ \beta_m \in \mathcal{B}. 
        \end{array}
\end{equation}
\subsection{Optimization Objective}
Assuming frame budget $\alpha_k$ and spatial budget $\beta_m$, we denote the inference accuracy achieved by the ViT model as $\phi(\alpha_k, \beta_m)$. The objective is to maximize the inference accuracy while adhering to a specified delay constraint. We formulate the optimization objective mathematically as follows:
\begin{subequations}
\begin{alignat}{2}
    \max_{\alpha_k, \beta_m} \quad & \phi(\alpha_k, \beta_m) &  \\
    \mathrm{s.t.} \quad & t_{\alpha_k,\beta_m} \le \delta, &\quad \forall  \alpha_k \in \mathcal{A} \ \textrm{and} \ \beta_m \in \mathcal{B} \,. 
\end{alignat}
\end{subequations}
where $\delta$ represents the maximum allowed completion time for a video recognition application.
\section{Design of STAE}
In this section, we first describe in detail the proposed FA module and SA module in the semantic encoder, as well as the FR module in the semantic decoder.
Subsequently, we introduce the training strategy of our proposed STAE.
\subsection{Spatiotemproal Attention Architecture}
The attention mechanism has been shown to improve the prediction performance of DNNs by enhancing the representation of features with more important information while suppressing unnecessary information interference \cite{Woo_2018_ECCV}. Inspired by this, we introduced FA and SA modules to evaluate the informative frames and pixels in each frame, respectively.
\subsubsection{Frame Attention Module}
The FA module focuses on which frames play an important role in determining the final inference results of the ViT model, i.e., it selects the frames that are decisive for prediction. As shown in Fig. 2, the input feature map $X$ is initially passed through both maximum pooling and average pooling operations along the frame dimension. The outputs of the pooling operations are then fed into an MLP layer, followed by element-wise summation and activation using a sigmoid function to generate the frame attention weight map $M_{\textit{FA}} \in \mathbb{R}^{F \times 1 \times 1 \times 1}$. The detailed process is described as
\begin{equation}
\small
M_{\textit{FA}} \hspace{-0.25mm}=\hspace{-0.25mm} \sigma \left(\textrm{MLP}\left(\textrm{Avgpool3D}\left(X \right)\right) +\textrm{MLP}\left(\textrm{Maxpool3D}\left(X \right)\right) \right),
\end{equation}
where $\sigma$ indicates the sigmoid function, $\textrm{MLP}$ represents the multilayer perceptron, and $\textrm{Avgpool3D}$ and $\textrm{Maxpool3D}$ denote the pooling operations. The frame attention weight map $M_{\textit{FA}}$ will be used to identify the informative frames tensor $X' \in \mathbb{R}^{\alpha_k \times C \times H \times W}$ based on the frame budget $\alpha_k$.
\subsubsection{Spatial Attention Module}
The SA module is designed to identify which pixels in an RGB image are most informative for making classification. This is achieved by using a similar mechanism as the FA module but applied at the pixel level within each frame, yielding a pixel-level attention weight map to represent the importance of each pixel in the frame. The implementation is illustrated in Fig. 2, where the selected informative frames tensor $X'$ is first subjected to a channel-based maximum and mean pooling operation, and then the two outputs are concatenated and passed through a $3 \times 3$ convolutional layer. Finally, the resulting feature map is activated using the sigmoid function to generate the spatial attention weight map $M_{\textit{SA}} \in \mathbb{R}^{\alpha_k \times 1 \times H \times W}$ given as
\begin{equation}
M_{\textit{SA}} = \sigma \left(\textrm{Conv}\left[\textrm{Avgpool3D}\left(X \right) ;\textrm{Maxpool3D}\left(X \right)\right] \right),
\end{equation} 
where $\textrm{Conv}$ is the convolutional operation. The spatial attention weight map $M_{\textit{SA}}$ is used to select the informative pixels based on the spatial budget $\beta_m$, resulting in the semantic tensor $X'' \in \mathbb{R}^{\alpha_k \times C \times \beta_m \times H \times W}$ for transmission.
\subsection{Feature Recovery Module} 
Due to the temporal correlation between different frames in a video, it is feasible to filling in missing or corrupted parts of one frame using information from the surrounding frames \cite{VideoInpaint}. As such, we design a light-weight FR module to recover missing regions in frames by jointly leveraging temporal and spatial details. As shown in Fig. 3, two sub-networks, i.e., a 3D CNN and a 3D-2D combined CNN, work together to implement the FR module. 

The entropy-decoded semantic tensor $X''$ is initially zero-filled to match the size of $X'$, resulting in the input feature map $X'_M \in \mathbb{R}^{\alpha_k \times C \times H \times W}$ of FR module. Meanwhile, a corresponding mask matrix $M$ is created, with ones and zeros indicating where pixels of $X'_M$ are available and where they are missing, respectively. The left sub-network utilizes 3D CNN to analyze the motion and dynamics of each frame thereby inferring the temporal-spatial structure of $X'_M$. Note that we use a down-sampled version of $X'_M$ and $M$, i.e., $\frac{H}{2} \times \frac{W}{2}$, as the input to reduce the huge computation overhead of 3D convolution, resulting in a low-resolution tensor $X'_{M,3D} \in \mathbb{R}^{\alpha_k \times C  \times \frac{H}{2} \times \frac{W}{2}}$. In this 3D CNN, we employ $3 \times 3$ kernels with a stride of $2$ to ensure that every pixel contributes and use the skip-connections as U-Net \cite{ronneberger2015u} to facilitate the feature mixture across different layers.
Another sub-network is a
3D-2D combined CNN, which applies 2D convolutional layers to perform completion frame by frame and thus generate the resulting complete frame $X'_R \in \mathbb{R}^{\alpha_k \times C  \times H \times W}$. The input of this 3D-2D combined CNN is the incomplete high-resolution frame of $X'_M$ with its mask matrix $M$. To tackle the issue that 2D CNN treats each frame independently without considering temporal coherence, we inject the information from the output of 3D CNN by extracting two feature maps of $X'_{M,3D}$ using two convolutional layers respectively. These two feature maps are added to the first and last layers of the same size in the 3D-2D combined CNN as temporal guidance.
\subsection{Training Strategy}
The proposed STAE can be trained in an end-to-end manner; however, it may take a long time to converge due to the model complexity and the multiple components that need to be fine-tuned. Therefore, we adopt a step-by-step training approach to achieve better performance and faster convergence as follows: 1) we first insert the FA module into the ViT model and train the resulting neural network to determine the importance of individual frames and identify the informative frames tensor $X'$ based on the frame budget $\alpha_k$;
2) we then freeze the FA module, feed the informative frames tensor $X'$ into the SA module, insert it into the ViT model, and train the resulting neural network to evaluate the importance of the pixels within each frame and generate the semantic tensor $X''$ based on the spatial budget $\beta_m$; 3) we finally freeze the FA and SA modules, feed the semantic tensor $X''$ into the FR module and insert it into the ViT model, and then fine-tune the resulting model to improve the inference accuracy. The detailed training process is described in Algorithm 1.
\section{Performance Evaluation}\label{simulation}
\subsection{Experimental Setup}
We consider the video action recognition task using the human motion dataset HMDB51 \cite{HMDB}, which divides 6849 clips into 51 action categories with at least 101 clips in each category. In this experiment, we consider an IoT device (Raspberry Pi 4B) and an ES (RTX 2080TI) to perform video action classification based on the popular ViT\_base model (patch size $16 \times 16 $, $12$ layers) with the input size of $16\times 3 \times 224\times224$ \cite{tong2022videomae}. To perform inference under time-varying wireless channel states, we select the semantic information with different frame budgets and spatial budgets to meet the deadline while ensuring high inference accuracy. The predefined frame budget and spatial budget are
$\alpha_k \in \mathcal{A} = \left\{1, 2,4, 8, 16\right\}$ and $\beta_m \in \mathcal{B} = \left\{100\%, 90\%, 80\%, 70\%, 60\%, 50\%, 40\%\right\}$ respectively.

To demonstrate the effectiveness of our proposed STAE architecture, we conduct a comparative analysis against the state-of-the-art baseline \cite{huang}, which proposed a deep learning-based image semantic coding (DeepISC) for feature extraction and recovery. Note that DeepISC only uses Huffman entropy coding to encode the latent representation but not the segmentation and the discrepancy between the input and synthesized data. To make a fair comparison, we incorporate our proposed FA module into DeepISC and evaluate its impact on the performance of the ViT model. For ease of presentation, we denote the methods used in this experiment as follows: 
\begin{itemize}
    \item ViT\_STAE: ViT model with STAE architecture;
    \item ViT\_DeepISC: ViT model with FA and DeepISC. 
\end{itemize}
\begin{figure}[]
    \centering
    \includegraphics[width=0.47\textwidth]{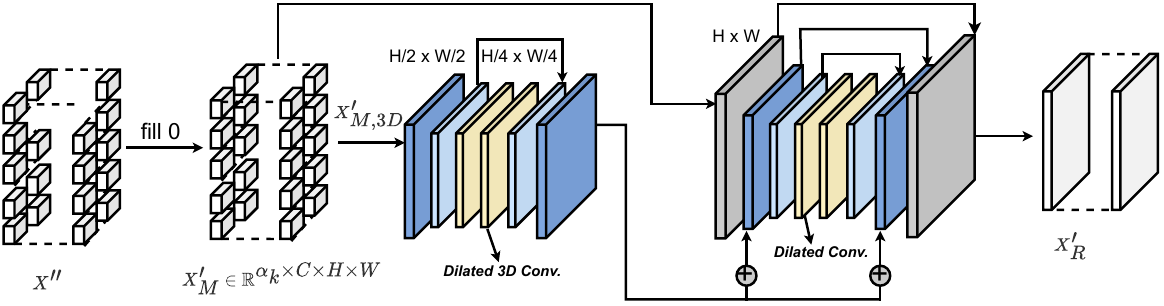}
    \caption{The feature recovery module of the Decoder.}
    \label{fig:decoder}
\end{figure}
\begin{algorithm}[t]
\caption{Training strategy for our proposed STAE}
\label{alg:example}
\begin{algorithmic}[1]
\REQUIRE Input training dataset $\mathcal{D}$, the set of frame budget $\mathcal{A}$ and spatial budget $\mathcal{B}$ and ViT model $\mathcal{G}$.
\ENSURE Output the set of STAE-enhanced ViT models $\mathbf{G} = \left\{\mathcal{G}_{\alpha_1}^{\beta_1}, \cdots, \mathcal{G}_{\alpha_1}^{\beta_M}, \cdots,\mathcal{G}_{\alpha_K}^{\beta_M}\right\}$.
\FOR{$k=1$ to $K$}
    \STATE Freeze ViT model and then insert FA module before ViT to result in a ViT\_FA model.
    \STATE Train ViT\_FA model on $X \in \mathcal{D}$ to generate $M_{\textit{fa}}$.
    \STATE Use $M_{\textit{fa}}$ to select informative tensor $X'$ based on frame budget $\alpha_k$.
    \FOR{$m=1$ to $M$}
    \STATE Freeze ViT\_FA model and insert SA module after FA module to generate a ViT\_FA+SA model.
    \STATE Feed $X'$ into ViT\_FA+SA from SA module and train the model to yield $M_{\textit{sa}}$.
    \STATE Use $M_{\textit{sa}}$ to select semantic tensor $X''$ based on spatial budget $\beta_m$.
    \STATE Freeze ViT\_FA+SA model and insert FR module after SA module to get ViT\_STAE model.
    \STATE Feed $X''$ into ViT\_STAE model from FR module and fine-tune model $\mathcal{G}_{\alpha_k}^{\beta_m}$ to improve accuracy.
    \ENDFOR
\ENDFOR
\RETURN $\mathbf{G}$
\end{algorithmic}
\end{algorithm}
\renewcommand\arraystretch{1.1}
\begin{table*}
    \begin{center}
    \caption{Inference accuracy and computation time (measured on Raspberry Pi 4B and RTX 2080TI)}
    \setlength{\tabcolsep}{0.95mm}{
    \begin{tabular}{cccccccccccccccccc}
        \Xhline{0.8 pt}
         \multirow{2}{*}{$\boldsymbol{\alpha_k}$} & \multirow{2}{*}{$t_{\alpha_k}^{\textit{ViT}}$ (ms)} & \multicolumn{3}{c}{$\boldsymbol{\beta_m=100\%}$ (VIT\_FA)} & \multicolumn{4}{c}{$\boldsymbol{\beta_m=40\%}$ (VIT\_FA+SA)} & \multicolumn{5}{c}{$\boldsymbol{\beta_m=40\%}$ (VIT\_STAE)} & \multicolumn{4}{c}{$\boldsymbol{\beta_m=40\%}$ (VIT\_DeepISC)} \\
         \cmidrule(lr){3-5} \cmidrule(lr){6-9} \cmidrule(lr){10-14} \cmidrule(lr){15-18}
             & & $S_{(\cdot)}$ (MB) & $\phi(\cdot)$ (\%)  & $t_{\alpha_k}^{\textit{FA}}$ & $S_{(\cdot)}$ & $\phi(\cdot)$  & $t_{\alpha_k}^{\textit{FA}}$ & $t_{(\cdot)}^{\textit{SA}}$ &$S_{(\cdot)}$ &$\phi(\cdot)$ & $t_{\alpha_k}^{\textit{FA}}$ & $t_{(\cdot)}^{\textit{SA}}$ & $t_{(\cdot)}^{\textit{FR}}$ &$S_{(\cdot)}$ & $\phi(\cdot)$ & $t_{\alpha_k}^{\textit{FA}}$ & $t_{(\cdot)}^{\textit{DeepISC}}$\\
        \Xhline{0.8 pt}
        \textbf{16}    &    $31.5$ &   $9.19$ &     $73.3(\pm0.24)$ &    $0$  &$3.67$  &  $71.7(\pm0.20)$ & $0$ & $43.3$ &$3.67$ &     $72.6(\pm0.18)$   & $0$ & $43.3$ &  $16.3$&$3.67$ &     $68.6(\pm0.14)$   & $0$ & $82.7$\\
        \textbf{8} &    $24.6$ &   $4.59$ &     $71.9(\pm0.21)$ &    $2.7$ &       $1.84$  &  $70.6(\pm0.22)$ &    $2.7$ & $36.1$ &       $1.84$ &     $71.3(\pm0.17)$   &    $2.7$ & $36.1$ &  $7.8$ &       $1.84$ &     $67.1(\pm0.19)$   &    $2.7$ & $67.3$  \\
        \textbf{4} &    $18.7$ &   $2.30$ &     $70.8(\pm0.20)$ &    $2.7$  &$0.92$  &  $69.5(\pm0.21)$ &    $2.7$ & $30.3$ &$0.92$ &     $70.1(\pm0.22)$   &    $2.7$ & $30.3$ &  $4.9$ &$0.92$ &     $66.0(\pm0.21)$   &    $2.7$ & $54.1$\\
        \textbf{2} &    $10.2$ &   $1.15$ &     $69.8(\pm0.19)$ &    $2.7$ & $0.46$  &  $68.5(\pm0.21)$ &    $2.7$ & $21.7$ & $0.46$ &     $69.1(\pm0.20)$  &    $2.7$ & $21.7$ &  $2.5$ & $0.46$ &     $64.9(\pm0.16)$  &    $2.7$ & $41.9$\\
        \textbf{1} &    $6.8$ &   $0.57$ &     $68.7(\pm0.21)$ &    $2.7$ & $0.23$  &  $67.6(\pm0.20)$ &    $2.7$ & $15.4$ & $0.23$ &    $68.1(\pm0.19)$ & $2.7$ & $15.4$  &  $1.6$ & $0.23$ &   $62.8(\pm0.19)$ & $2.7$ & $31.5$\\
        \Xhline{0.8 pt}
    \end{tabular} }
    \label{tab:acc}
    \end{center}
        \vspace{-4mm}
\end{table*}
\renewcommand\arraystretch{1.1}
\begin{table*}[]
    \begin{center}
    \caption{Entropy encoding time (measured on Raspberry Pi 4B) and decoding time (measured on RTX 2080TI)}
    \setlength{\tabcolsep}{1.2mm}{
    \begin{tabular}{ccccccccccccc}
        \Xhline{0.8pt}
         \multirow{2}{*}{$\boldsymbol{\alpha_k}$} &\multicolumn{3}{c}{$\boldsymbol{\beta_m = 100\%}$ (VIT\_STAE)} &\multicolumn{3}{c}{$\boldsymbol{\beta_m = 100\%}$ (VIT\_DeepISC)} &\multicolumn{3}{c}{$\boldsymbol{\beta_m = 40\%}$ (VIT\_STAE)}&\multicolumn{3}{c}{$\boldsymbol{\beta_m = 40\%}$ (VIT\_DeepISC)}\\
         \cmidrule(lr){2-4} \cmidrule(lr){5-7} \cmidrule(lr){8-10} \cmidrule(lr){11-13} 
              & Entropy  & $t_{\alpha_k,\beta_m}^{\textit{EE}}$(ms)  & $t_{\alpha_k,\beta_m}^{\textit{ED}}$ & Entropy  & $t_{\alpha_k,\beta_m}^{\textit{EE}}$   & $t_{\alpha_k,\beta_m}^{\textit{ED}}$  & Entropy  & $t_{\alpha_k,\beta_m}^{\textit{EE}}$   & $t_{\alpha_k,\beta_m}^{\textit{ED}}$ & Entropy  & $t_{\alpha_k,\beta_m}^{\textit{EE}}$   & $t_{\alpha_k,\beta_m}^{\textit{ED}}$ \\
        \Xhline{0.8pt}
        \textbf{16}  &$15.37(\pm0.14)$ &10.53 &8.63 &$14.72(\pm0.19)$ &8.73 &7.36 &$14.03(\pm0.12)$ &9.35 &7.92 &$13.92(\pm0.19)$ &8.31 &7.12 \\
        \textbf{8} &$14.48(\pm0.16)$ &8.25 &7.41 &$13.71(\pm0.14)$ &7.42 &5.83 &$13.01(\pm0.15)$ &7.95 &6.41 &$12.94(\pm0.12)$ &7.21 &5.43\\
        \textbf{4} &$13.86(\pm0.12)$ &6.94 &5.52 &$13.27(\pm0.15)$ &5.49 &4.25 &$12.89(\pm0.17)$ &5.90 &4.91 &$12.47(\pm0.14)$ &5.08 &4.07 \\
        \textbf{2} &$13.13(\pm0.20)$ &4.76 &3.37 &$12.91(\pm0.11)$ &4.06 &2.79 &$12.73(\pm0.16)$ &4.24 &3.02 &$12.42(\pm0.19)$ &3.89 &2.54\\
        \textbf{1} &$12.62(\pm0.19)$ &3.03 &2.04 &$12.33(\pm0.16)$ &2.30 &1.47 &$12.26(\pm0.18)$ &2.79 &1.76 &$12.11(\pm0.12)$ &2.02 &1.35 \\
        \Xhline{0.8pt}
    \end{tabular}}
    \label{tab:entropy}
    \end{center}
        \vspace{-6mm}
\end{table*}
\begin{figure}
    \centering
\includegraphics[width=0.33\textwidth]{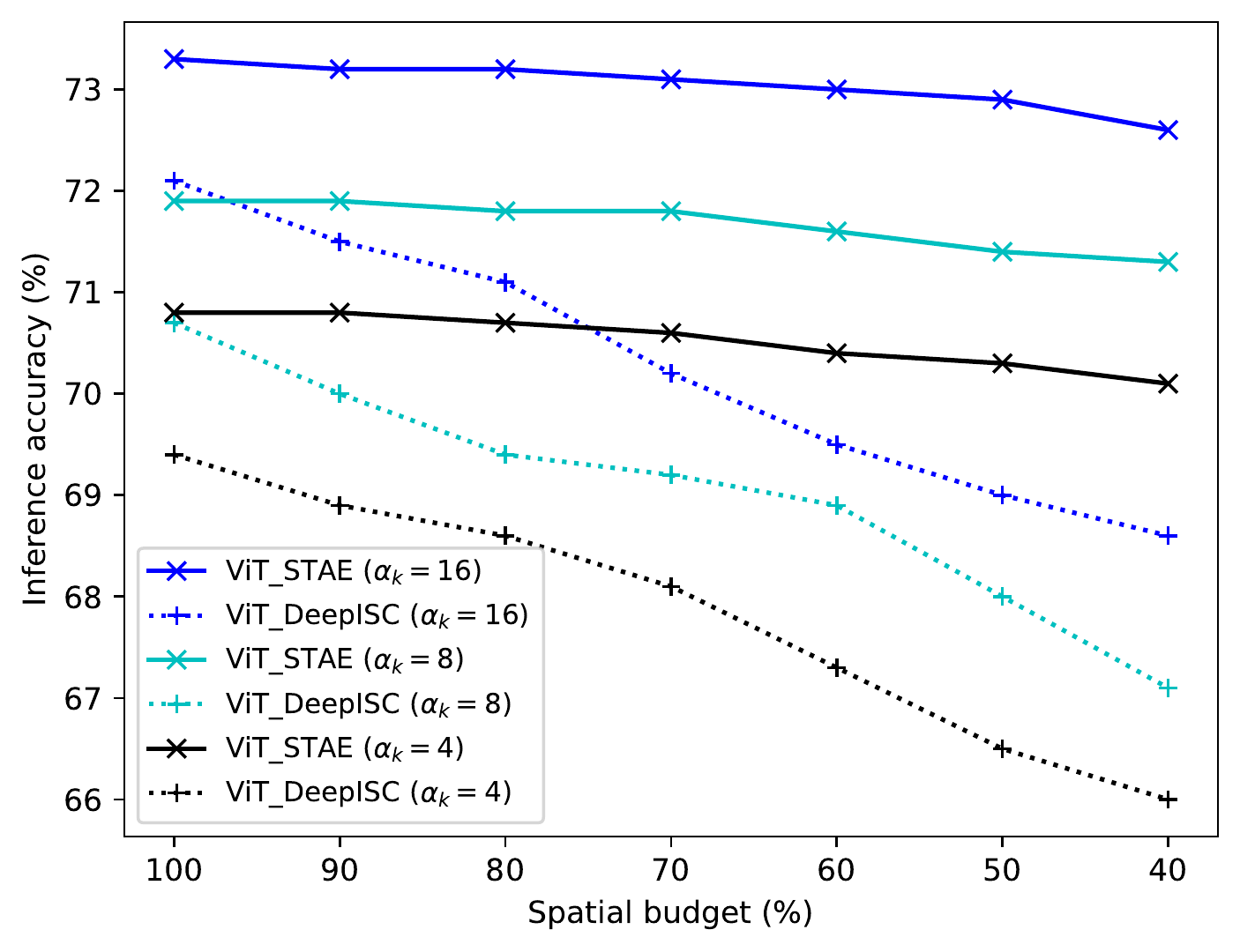}
    \caption{Accuracy under different frame and spatial budgets.}
        \vspace{-5mm}
    \label{fig:acc}
\end{figure}
\begin{figure*}
    \centering
    \subfigure[Total completion time under different data rates]{
        \label{fig:time1}    \includegraphics[width=0.33\textwidth]{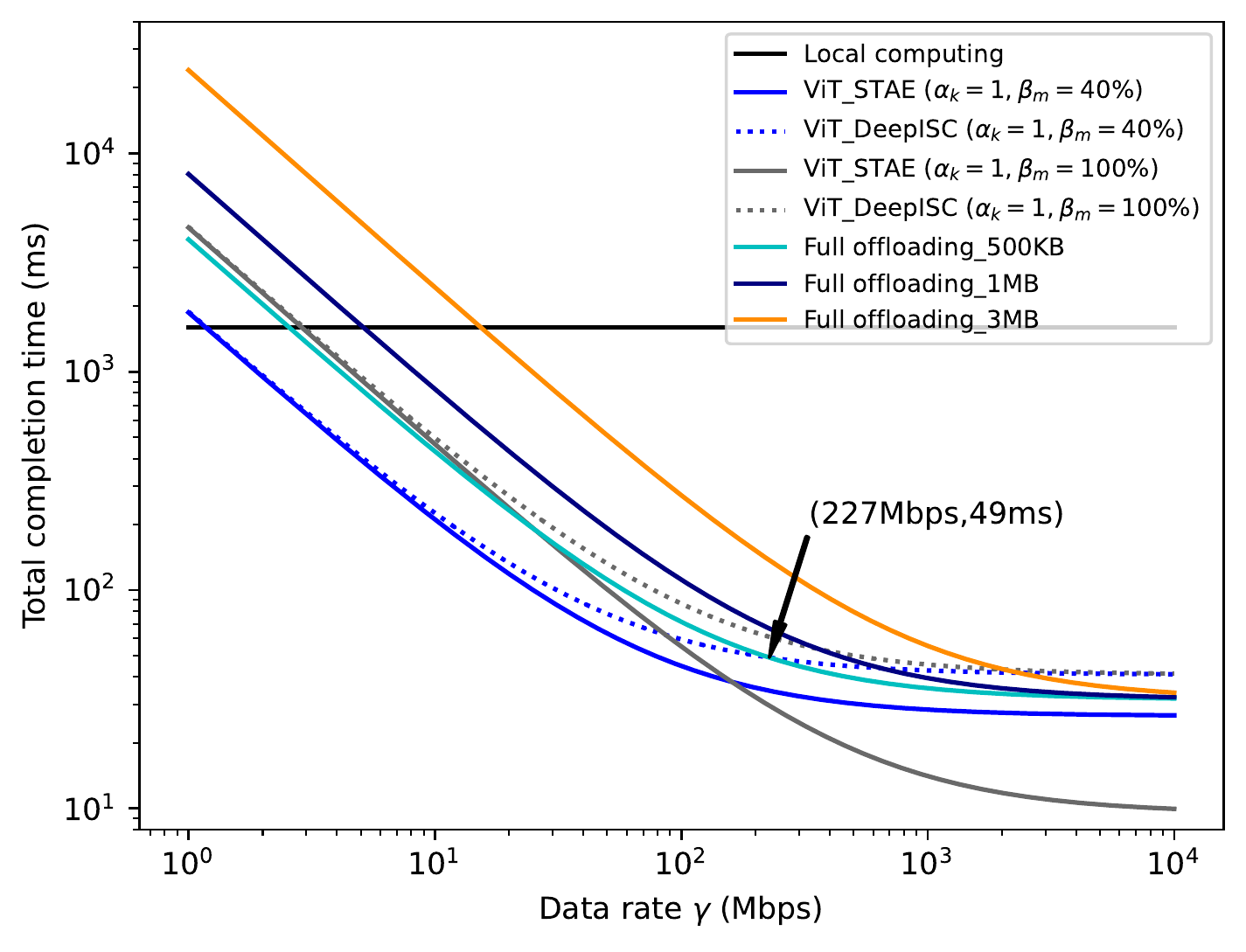}}
    \hspace{15mm}
    \subfigure[Inference accuracy vs. latency]{
    \includegraphics[width=0.33\textwidth]{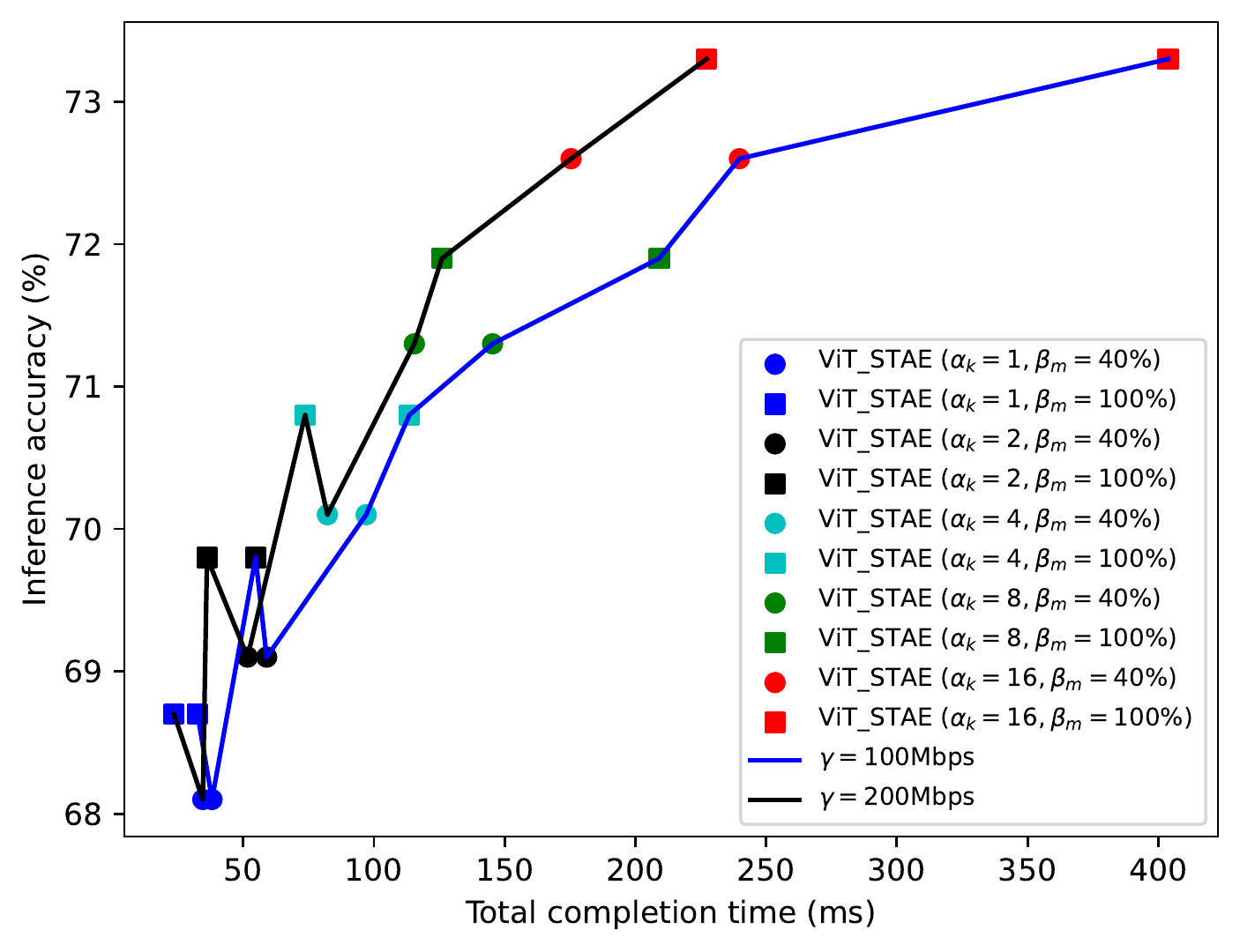}
        \label{fig:time2}}
        \vspace{-2mm}
    \caption{Tradeoff between inference accuracy and latency under different transmission data rates}
    \vspace{-6mm}
\end{figure*}
\subsection{Performance Evaluation and Analysis}
Since the importance of different pixels in each frame is not measured in DeepISC, we assume the pixels are randomly selected based on the spatial budget $\beta_m$. Additionally, ViT\_STAE uses the SA module and FR module to compress and recover the data only if the spatial budget $\beta_m \neq 100\%$.
\addtolength{\topmargin}{0.05in}
\subsubsection{ \textbf{Inference Accuracy under Different Frame and Spatial Budgets}}
Fig. \ref{fig:acc} compares the inference accuracy of the proposed ViT\_STAE and ViT\_DeepISC under different frame and spatial budgets. It is clear that the accuracy of both methods decreases as the frame and spatial budgets decrease because the lack of crucial information for inference often leads to inaccurate predictions. In addition, our proposed ViT\_STAE shows that compression in the spatial dimension achieves higher accuracy than compression in the frame dimension with the same communication overhead.
For instance, the accuracy at $\alpha_k=16$ and $\beta_m=50\%$ is superior to that at $\alpha_k=8$ and $\beta_m=100\%$. This is because frames often contain extraneous background information that is not useful for prediction; however, selecting fewer frames may miss important semantic information.
Furthermore, our proposed ViT\_STAE outperforms ViT\_DeepISC at any spatial budget for a given frame budget. This is due to the fact that ViT\_STAE is capable of evaluating the importance of pixels and transmitting the most critical semantic information. In contrast, ViT\_DeepISC needs to send not only the latent representation but also the semantic segmentation and the differences between the input and synthesized data, therefore it may miss critical semantic information by randomly transmitting the same amount of spatial information.

Table \ref{tab:acc} lists the inference accuracy and computation time of ViT\_STAE and ViT\_DeepISC under various frame and spatial budgets, where we use ViT\_FA and  ViT\_FA+SA to denote the ViT model with FA module and the ViT model with FA and SA module, respectively. The results indicate that ViT\_STAE outperforms ViT\_FA+SA in terms of accuracy, which demonstrates the effectiveness of our proposed FR module. Compared to ViT\_DeepISC, our proposed ViT\_STAE achieves superior inference accuracy of up to $5.3\%$ (i.e., $68.1\%-61.8\%$) with reduced computation time and the same transmission data size. Furthermore, we present the entropy encoding and decoding time of ViT\_STAE and ViT\_DeepISC in Table \ref{tab:entropy}. Note that in ViT\_DeepISC, entropy coding is applied only for the latent representation but not for all the data.
Together with entropy coding, our proposed ViT\_STAE can compress the video data up to more than $104 \times$ (i.e., $16/1 \times 100/40 \times 32/12.26$) at $\alpha_k = 1$ and $\beta_m = 40\%$, with an accuracy loss of only about $5.2\%$ (i.e., $73.3\%-68.1\%$). 
\subsubsection{ \textbf{Tradeoff between Inference Accuracy and Latency under Time-varying Wireless Channel}}
Since offloading channel state is stochastic in practice, we compare the task completion time of VIT\_STAE and ViT\_DeepISC at various transmission data rates $\gamma$. In Fig. \ref{fig:time1}, we plot the completion time of the methods at $\alpha_k =1$ and $\beta_m = 40\%$. Note that we only use entropy coding when the introduced encoding-decoding time is less than the reduced communication time.
Noticeably, the completion time of local computing remains constant independent of the transmission data rate and the size of the input video, and the communication time of VIT\_STAE and ViT\_DeepISC is not affected by the size of the input video. We can see that local computing is beneficial only when the transmission rate is below 1.2 Mbps. In addition, VIT\_STAE can perform inference faster than ViT\_DeepISC and full offload, while ViT\_DeepISC is superior to full offloading only if the transmission data rate is below 227 Mbps. 

In general, high compression can accelerate inference but at the cost of accuracy loss. In Fig. \ref{fig:acc} and Table \ref{tab:acc}, we see that the SA and FR modules introduced by VIT\_STAE do not yield much improvement in inference accuracy but result in significant computational overhead compared to VIT\_FA. For example, VIT\_STAE achieves less than 1\% gain of accuracy at $\alpha_k = 16$ and $\beta_m = 50\%$ than at $\alpha_k = 8$ and $\beta_m = 100\%$; however, the introduced computation time is $59.6$ ms (i.e., $43.3+16.3$) at the same amount of transmission data. As such, the reduced communication time of VIT\_STAE might be able to pay off the introduced computation time only if $\beta_m < 50\%$ compared to VIT\_FA. To trade off latency and accuracy for time-constrained IoT applications, we can select the optimal frame budget and spatial budget to achieve the best possible accuracy within the time constraint. Fig.~\ref{fig:time2} shows the inference accuracy and total completion time at different frame budgets, spatial budgets and data rates. We can see that spatial attention can provide feasible solutions in some circumstances but not always. For example, at the data rate of 100 Mbps and latency constraint of 100 ms, $\alpha_k = 4, \beta_m = 40\%$ is the best feasible option, however, at 200 Mbps and the same time constraint, $\alpha_k = 4, \beta_m = 100\%$ becomes the best feasible option and $\alpha_k = 4, \beta_m = 40\%$ does not satisfy any latency constraint. 
\section{Conclusion}\label{conclusion}
This paper presents a novel STAE architecture for video action recognition in MEC scenarios. By incorporating an FA and SA module in the semantic encoder, the proposed STAE efficiently learns informative video representations by identifying the most important frames and pixels in each frame. In addition, entropy coding is used to remove statistical redundancy in the semantically compressed data to further reduce communication overhead. To improve inference accuracy, a lightweight semantic decoder is designed to reconstruct missing information by using a 3D-2D CNN combined architecture to simultaneously learn temporal and spatial information from the received data. Experimental results show that ViT\_STAE achieves $104 \times$ better in compressing HMDB51 with only 5\% accuracy loss while being significantly faster and more efficient than the existing method ViT\_DeepISC. The proposed approach can serve as a foundation for our future research in MEC and video analysis. 
\section*{Acknowledgment}
This work is supported by Agile-IoT project (Grant No. 9131-00119B) granted by the Danish Council for Independent Research.

\bibliographystyle{IEEEtran}
\bibliography{main}
\end{document}